\documentclass[11pt]{article}

\usepackage[preprint]{acl}
\usepackage{booktabs}
\usepackage{hyperref}
\usepackage{url}
\usepackage[utf8]{inputenc}
\usepackage{graphicx}
\usepackage{xcolor}
\usepackage[normalem]{ulem}
\usepackage{times}
\usepackage{latexsym}
\usepackage{amsmath}
\usepackage{float}
\usepackage[T1]{fontenc}
\newcommand{\tokU}[1]{\texttt{\uline{#1}}} 
\newcommand{\tokB}[1]{\texttt{\textbf{#1}}} 
\newcommand{\tokR}[1]{\texttt{\textcolor{red}{#1}}} 

\usepackage[utf8]{inputenc}

\usepackage{microtype}

\usepackage{inconsolata}

\usepackage{graphicx}

\title{How Do Latent Reasoning Methods Perform Under Weak and Strong Supervision?}

\author{
 \textbf{Yingqian Cui\textsuperscript{1,2}\thanks{{Work done during her internship at Amazon.}}},
  \textbf{Zhenwei Dai\textsuperscript{1}},
 \textbf{Bing He\textsuperscript{1}},
  \textbf{Zhan Shi\textsuperscript{1}},
 \textbf{Hui Liu\textsuperscript{1}},
 \textbf{Rui Sun\textsuperscript{1}},\\
  \textbf{Zhiji Liu\textsuperscript{1}},
 \textbf{Yue Xing\textsuperscript{2}},
 \textbf{Jiliang Tang\textsuperscript{2}},
 \textbf{Benoit Dumoulin \textsuperscript{1}}
\\
\\
 \textsuperscript{1}Amazon,
 \textsuperscript{2}Michigan State University
\\
}

\begin{document}
\maketitle
\begin{abstract}
Latent reasoning has been recently proposed as a reasoning paradigm and performs multi-step reasoning through generating steps in the latent space instead of the textual space. This paradigm enables reasoning beyond discrete language tokens by performing multi-step computation in continuous latent spaces. Although there have been numerous studies focusing on improving the performance of latent reasoning, its internal mechanisms remain not fully investigated. In this work, we conduct a comprehensive analysis of latent reasoning methods to better understand the role and behavior of latent representation in the process. We identify two key issues across latent reasoning methods with different levels of supervision. First, we observe pervasive shortcut behavior, where they achieve high accuracy without relying on latent reasoning. Second, we examine the hypothesis that latent reasoning supports BFS-like exploration in latent space, and find that while latent representations can encode multiple possibilities, the reasoning process does not faithfully implement structured search, but instead exhibits implicit pruning and compression. Finally, our findings reveal a trade-off associated with supervision strength: stronger supervision mitigates shortcut behavior but restricts the ability of latent representations to maintain diverse hypotheses, whereas weaker supervision allows richer latent representations at the cost of increased shortcut behavior. 
\end{abstract}

\section{Introduction}

hain-of-thought (CoT) prompting has emerged as a powerful technique for enhancing the reasoning capabilities of Large Language Models (LLMs). By explicitly generating intermediate steps, CoT enables models to decompose complex problems into fine-grained units and achieve substantial performance gains across a wide range of reasoning tasks~\citep{wei2022chain,yao2023react}.

{Despite its effectiveness, CoT reasoning is constrained by its reliance on natural language, and latent reasoning has been proposed to overcome the limitation:} Textual reasoning must be expressed as sequences of discrete tokens, potentially limiting the expressiveness of the underlying representations and biasing models toward linguistically convenient reasoning patterns~\citep{hao2024training, deng2024explicit}. In implicit or latent reasoning, models perform iterative computation in hidden space without explicitly generating intermediate reasoning steps~\citep{hao2024training,deng2024explicit,liu2024expediting,shen2025CODI,tan2025think}. Such approaches aim to explore the potential of reasoning beyond language, enabling models to leverage continuous, high-dimensional representations that support richer forms of reasoning. 

While extensive follow-up works continue to improve the performance of latent reasoning methods, the fundamental nature of latent reasoning remains not fully understood. Compared to explicit reasoning, which exposes intermediate steps in text, latent reasoning operates largely as a black-box process in which the intermediate reasoning states are not directly observable or interpretable, making it difficult to determine whether models are engaging in genuine multi-step reasoning.
{Besides, while Coconut~\citep{hao2024training} argues that iterative latent reasoning may enable the model to explore multiple possible reasoning paths before committing to an output and conjectures that this process resembles parallel breadth-first search (BFS) in the latent space, it remains unclear whether this conjecture holds consistently across different tasks or generalizes to other latent reasoning methods or not.
}
Without a clearer understanding of what intermediate latent steps represent, it is difficult to assess the robustness, generalization, and interpretability of these methods.

In this work, we analyze the internal mechanisms of latent reasoning. To begin with, from a training perspective, we categorize existing latent reasoning methods into two classes based on their supervision signals during training: (i) methods relying on weak or outcome-level supervision, where latent generation is largely unconstrained during training (\textit{weak supervision})~\citep{hao2024training,shen2025CODI}, and (ii) methods trained with explicit, fine-grained supervision that aligns latent states with intermediate reasoning steps (\textit{strong supervision})~\citep{wei2025sim,tan2025think}.

{We conduct a series of comprehensive experiments to examine the behavior of existing latent reasoning methods:}





First, we examine how reasoning performance evolves as a function of latent step depth (the number of consecutive latent steps executed
before final answer generation). Surprisingly, we find that in some specific tasks, the accuracy of some methods remains very high even when latent reasoning is entirely disabled (i.e., depth = 0), suggesting that models may bypass multi-step reasoning and rely on alternative shortcuts. We then provide additional evidence of this shortcut reliance through interventional evaluations and attention-based analyses.
Notably, while the shortcut issues have been mentioned by previous literature~\citep{yu2024llms,zhang2025latent}, our work provides a more comprehensive perspective by systematically examining how different training supervision schemes influence shortcut behavior, and what kind of tasks are more vulnerable to such issues.

 
Second, we revisit the parallel BFS hypothesis {by \citet{hao2024training}}. 
We find that while a single latent representation can indeed encode multiple candidates for a reasoning step, the overall reasoning process does not strictly follow a BFS pattern. Instead, latent exploration exhibits implicit pruning behaviors, whereby certain candidate paths are suppressed before sufficient exploration has been carried out as latent iterations progress. Moreover, even when a latent can encode a richer set of possibilities, the final prediction does not consistently concentrate probability mass on the correct solution. {This suggests that, while latent reasoning achieves compression by aligning explicit reasoning steps with latent states during training, its effect may be largely limited to reducing the number of generation steps, rather than realizing genuine BFS-style exploration.}


Finally, we also reveal a trade-off governed by supervision strength: Stronger supervision stabilizes reasoning and reduces shortcut behavior, but limits the diversity of candidate trajectories maintained in latent space. 

Taken together, this study provides new insights into the internal mechanisms of implicit reasoning models. By revealing how latent generation departs from ideal searching behaviors, our analysis offers insights into the future design of more capable and robust latent reasoning systems.

\section{Background}
In this section, we provide background for our analysis by introducing the formulation of latent reasoning and a categorization of existing methods based on the design of their training scheme.

\subsection{Latent Reasoning Mechanism.} 
Latent (or implicit) reasoning performs intermediate reasoning entirely in the latent space, without explicitly decoding intermediate steps into natural language. Instead of generating and appending reasoning tokens, the model performs reasoning by recursively using the latent state from the previous iteration as the input embedding to the next, and returns to text space only at the final step to generate the answer~\citep{hao2024training, shen2025CODI, wei2025sim}.

Specifically, let $x={(x_1, x_2,...,x_n)}$ denote the input token sequence, $E(x)$ be the token embedding of $x$ and $F_{\theta}$ be the transformer forward function. In \textbf{\textit{standard autoregressive generation}}, a transformer computes latent states $h_t = F_\theta\!\big(E(x_1),\dots,E(x_t)\big)$,
and predicts the next token distribution by $p(x_{t+1}\mid x_{\le t})=\text{Softmax}(W h_t).$

In contrast, \textbf{\emph{latent} (or \emph{implicit}) reasoning} replaces the discrete token-level feedback loop with a continuous latent one. Let $c_t$ denote the continuous thought at step $t$ (typically $c_t = h_t$ or a projection of $h_t$).
Rather than sampling a token and feeding back its embedding $E(x_{t+1})$, the model directly reuses the latent state from the previous iteration as the input embedding for the next step:
\[
e_{t+1} \leftarrow c_t,
\quad
h_{t+1} = F_\theta\!\big(E(x_1), \dots, E(x_t), e_{t+1}\big).
\]
After $T$ consecutive latent steps, the model returns to text space and generates the final answer tokens via the standard output head. Specifically, denoting $y={(y_1, y_2,...,y_m)}$ to be the output token sequence, we have  
{\small\[ y_{1} = \text{Softmax}(Wh_{t+T}), \quad y_{k+1} = \text{Softmax}(Wh_{k}), \quad \text{where}\]
\[h_{k} = F_\theta\!\big(E(x_1), \dots, E(x_t), e_{t+1}\dots e_{t+T}, E(y_1),\dots, E(y_k )\big).
\]}
Intuitively, this enables the model to ``think'' in a continuous latent space prior to textual generation. 
According to~\citet{hao2024training, zhu2025reasoning}, as intermediate states are not collapsed into deterministic discrete tokens, a single latent representation can, in principle, encode multiple trajectories simultaneously. Iteratively generating such latent steps has therefore been hypothesized to resemble parallel breadth-first exploration, where multiple candidate reasoning paths are implicitly maintained before final textual decoding.

\subsection{Weak/Strong Supervision Training Scheme.} \label{sec:back_sup}

Based on the training objectives and supervision granularity, we categorize the existing latent reasoning methods into two groups: approaches with \textbf{\emph{weak supervision}} and \textbf{\emph{strong supervision}}. Specifically, methods with weak supervision rely primarily on outcome-level objectives: latent reasoning is not directly constrained, but is learned only through its effects on the follow-up steps and the correctness of the final answer. In contrast, strongly supervised methods impose explicit and fine-grained supervision on latent representations, often through decoder-based objectives or token-level compression mechanisms. {Given the above definition, we categorize the four representative methods considered in this paper:}

\textbf{Coconut}~\citep{hao2024training} adopts a stage-wise training scheme that progressively shifts reasoning from text space to the latent space. Across training stages, the model increases the number of latent reasoning steps while correspondingly reducing the number of explicit textual reasoning steps. At each stage, supervision is provided through cross-entropy loss on the remaining textual predictions, where the model is trained to match the ground-truth in text space.

\textbf{CODI}~\citep{shen2025CODI} employs a unified training scheme without explicit stages. Its objective consists of two complementary loss terms. The first term supervises the learning of the latent steps using the final textual solution as an outcome-level objective. The second term introduces a teacher--student distillation objective, where latent representations produced by a teacher model after textual reasoning are used to supervise the student’s final latent states.

\textbf{SIM-CoT}~\citep{wei2025sim} extends the latent reasoning frameworks of Coconut and CODI by introducing an additional explanation loss. Specifically, it applies the language model decoder to the intermediate latent states and optimizes a reconstruction loss that conditions on each latent state to recover the corresponding textual reasoning step, which explicitly provides stronger supervision over reasoning dynamics.

\textbf{CoLaR}~\citep{tan2025think} applies supervision through token-level compression. Given a compression factor $c$, CoLaR averages the latent representations of every $c$ consecutive tokens and uses the resulting compressed representations as the supervision targets during training. In addition, standard supervision on the final textual answer is applied. This compression-based supervision directly constrains latent representations to align with aggregated token-level information.

Among the methods, Coconut and CODI rely on relatively weak supervision, as latent states are trained indirectly via final outputs or final latent representation alignment, whereas SIM-CoT and CoLaR impose stronger supervision through explicit alignment with intermediate textual steps or token-level representations.






\section{Other Related Works}
\paragraph{Additional implicit reasoning methods} In addition to the four representative methods that our analysis focuses on, there exist many other works that propose effective implicit reasoning approaches. For example, \citet{cheng2024compressed} introduce CCoT, which utilizes continuous, variable-length contemplation tokens to conduct implicit reasoning. \citet{liu2024expediting} propose HCoT, which applies a two-stage framework for achieving latent representation and final answer alignment. Another stream of studies, including CoCoMix~\citep{tack2025llm} and PonderLM-2~\citep{zeng2025ponderlm}, investigates pretraining strategies for continuous latent reasoning. Additionally, several recent works~\citep{geiping2025scaling, saunshi2025reasoning,wang2025hierarchical,zhu2025scaling} have explored recurrent or looped transformer architectures to generate continuous thoughts that simulate CoT reasoning in latent states. Other follow-up works have investigated test-time scaling approaches for latent reasoning~\citep{you2025parallel, ye2025thinking}.

\paragraph{Analytical work for latent reasoning methods} 
Beyond proposing new methods, a few recent works have also conducted theoretical or empirical analyses to better understand the mechanisms underlying latent reasoning.

Theoretically, \citet{zhu2025reasoning} leverage a graph reasoning problem to prove the effectiveness of latent reasoning with a simplified transformer structure and demonstrate that each continuous thought encodes multiple searches simultaneously. \citet{zhu2025emergence} further study the training dynamics of continuous reasoning and explains how the superposition of reasoning traces emerges in gradient-based training.
\citet{xu2025formal} indicate that while latent reasoning enables efficient parallel computation in latent space, explicit reasoning supports approximate counting and sampling via stochastic decoding. Although these theoretical works provide support for the hypothesis about the parallel search in latent reasoning, their analyses are largely conducted under simplified settings, and it remains unclear to what extent the proposed parallel-search behavior emerges in practical latent reasoning systems trained on real-world data.

Empirically, \citet{yu2024llms,zhang2025latent} provide analysis on whether latent reasoning generates effective intermediate reasoning steps. \citet{yu2024llms} shows that prompt-based latent reasoning often fails to induce genuine internal reasoning, while training-based approaches can produce more meaningful latent steps. \citet{zhang2025latent} indicate that Coconut exhibits shortcut dependence rather than faithfully executing a reasoning process in certain datasets.
While our findings partially overlap with those of these prior studies, we provide a more comprehensive analysis by systematically examining how different training schemes influence the shortcut behavior and what kind of tasks are more vulnerable to such issues.

\section{Shortcut Behavior of Latent Reasoning}
In this section, we present an analysis of multiple latent reasoning models to reveal varying degrees of collapse behavior. We begin by introducing the experimental setup in Section~\ref{sec:exp_set}. We then investigate the dynamics of
latent reasoning in Section~\ref{sec:length} by examining how reasoning performance varies with latent step depth at inference time. Building on these observations, Section~\ref{sec:shortcut} further investigates the shortcut behaviors in latent reasoning models through interventional analysis and investigation with attention score.


\subsection{Experimental Settings}
\label{sec:exp_set}

\paragraph{Latent Reasoning Methods.}
Our investigation involves the aforementioned four methods, i.e., Coconut~\citep{hao2024training}, CODI~\citep{shen2025CODI}, SIM-CoT~\cite{wei2025sim}, and CoLaR~\cite{tan2025think}.
Notably, SIM-CoT can be applied on top of both Coconut and CODI. We report results for the better-performing variant, SIM-CoT+CODI, and refer to it as SIM-CoT throughout the remaining part of the paper. In addition, CoLaR allows varying compression factors. To make the number of latent steps comparable with other methods, we set the compression rate to $5$ for CoLaR in all experiments.
As mentioned in Section~\ref{sec:back_sup}, we categorize Coconut and CODI as methods with \emph{Weak Supervision}, and SIM-CoT and CoLaR as methods with \emph{Strong Supervision}.

\paragraph{Base models and Benchmarks.}
All latent reasoning methods in our experiments are trained on two backbone LMs: GPT-2~\citep{radford2019language} and Llama-3.2-1B-Instruct~\citep{dubey2024llama}.
We use full fine-tuning for GPT-2 and LoRA fine-tuning~\citep{hu2022lora} for Llama-3.2-1B-Instruct. Additional training details and hyperparameter settings are provided in Appendix~\ref{appd:exp_setup}.
Our experiments are based on two reasoning datasets: GSM8K-Aug~\citep{deng2024explicit}, an augmented version of GSM8k~\citep{cobbe2021training}, which is a grade-school math word problem benchmark, and ProsQA~\citep{hao2024training}, a more complicated version of ProntoQA~\citep{saparov2022language}, which requires multi-step first-order logical reasoning over compositional rules.

\subsection{Influence of the latent length}
\label{sec:length}
To better understand the nature of reasoning information encoded in latent representations, we begin by analyzing how latent steps evolve throughout reasoning by measuring how reasoning accuracy changes across different numbers of latent steps. 

\begin{figure*}[h]
    \centering
    \includegraphics[width=1.01\textwidth]{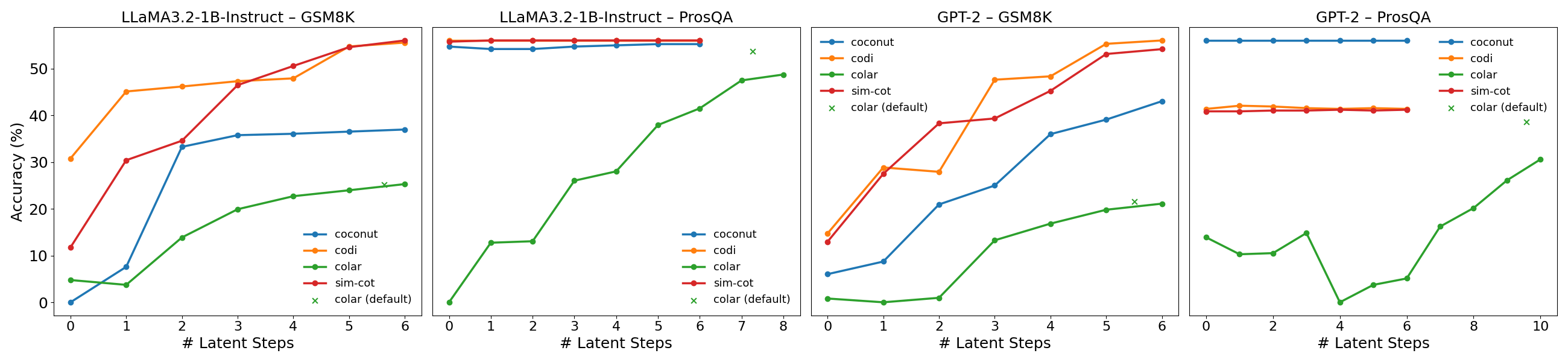}
    \caption{Performance of different methods across different numbers of latent steps}
\label{fig:length}
\end{figure*}

We train all methods using their default training configurations, with a fixed latent length for all methods except CoLaR. During inference, we manually vary the number of latent steps to study its effect on performance.
For CoLaR, the default training and inference process allows different examples to use different latent lengths. For controlled comparison, we modify the inference procedure by fixing the latent length and forcing the model to produce the final answer by explicitly inserting the token that triggers final-answer generation. We also report the results under the default setting by plotting the average accuracy against the average number of latent steps used in the test set. 
The results are shown in Figure~\ref{fig:length}. For both datasets, we observe a shortcut reliance phenomenon:

First, for GSM8K, when the number of latent steps increases, the accuracy in general increases for all methods. This is intuitive, as GSM8K requires multi-step numerical reasoning, and a sufficient number of latent steps allows models to carry out more complete step-by-step reasoning. However, when the latent depth is set to zero, accuracy does not typically collapse to zero. Notably, CODI retains around 30\% accuracy on GSM8K when using LLaMA, suggesting that a nontrivial portion of the performance arises from direct answer generation, rather than based on stepwise reasoning.

Second, on the ProsQA dataset, most methods (except CoLaR) exhibit almost no performance change as the number of latent steps changes, even when the latent length is reduced to zero. This abnormal behavior suggests that, for ProsQA, these methods do not strongly rely on latent reasoning steps to achieve high accuracy, but captures a shortcut to answer the question instead. To explain this, compared to GSM8K, ProsQA involves simpler logical structures and more repetitive reasoning patterns, making it easier for models to exploit surface-level cues without performing genuine multi-step reasoning, which amplifies the shortcut behavior.


Finally, CoLaR appears to be the most sensitive method to the number of latent steps. When the latent length is reduced to zero, performance on GSM8K drops to nearly zero, and on ProsQA it falls close to random-guess performance (approximately 50\% for this binary task). This pattern indicates that CoLaR relies heavily on latent representations to perform reasoning across both datasets. This is closely related to CoLaR’s training process, which applies strong supervision by aligning latent representations with token-level information. With stricter learning signals beyond the final answer, the model is forced to encode meaningful information in intermediate reasoning steps, making it less likely to learn shortcut mappings.


\subsection{Interventional Analysis}
\label{sec:shortcut}

In the previous subsection, we observed indications of a potential shortcut behavior. To further validate this hypothesis, in this subsection, we conduct an interventional analysis by applying controlled noise perturbations to the latent steps to assess the model’s reliance on latent reasoning.

Specifically, after latent reasoning terminates and the model signals the
transition to final answer generation, we inject a random noise sampled from a
multivariate Gaussian distribution,
$\boldsymbol{\epsilon} \sim \mathcal{N}(\mathbf{0}, \Sigma)$, into the
embedding corresponding to the latent token. The model then proceeds to generate the final answer.
To ensure the perturbation is strong enough to corrupt latent representations,
the injected noise ($\Sigma=\sigma^2\mathbf{I}$, $\sigma=100$) is
much larger in magnitude than the original latent embeddings (the average $\ell_2$ norms of the latent embeddings are $24.42$ for GPT-2 and $44.08$  for LLaMA3.2-1B, with embedding dimensions of 768 and 2048, respectively).


Table~\ref{tab:noise} reports the accuracy of different methods before and after noise injection. For a direct comparison, we apply the same intervention to explicit reasoning by injecting noise into the embedding of the CoT token immediately before the final answer is generated (denoted as Standard CoT). 

From the results in the table, we observe a clear contrast between standard CoT and latent CoT. When noise is injected into the reasoning token, the accuracy of standard CoT drops to nearly zero. In contrast, almost all latent methods retain non-zero performance under strong noise perturbations that disrupt latent representations, indicating the presence of shortcut behaviors.

The emergence of shortcut behavior is largely consistent with the trends observed in our analysis of latent step depth. On the simpler ProsQA dataset, only CoLaR, the method with stronger supervision, shows a substantial performance degradation under noise injection, while other latent methods remain largely unaffected. On GSM8K, while all latent methods experience notable
performance drops, they retain non-trivial accuracy, with the effect being most pronounced for Coconut and CODI, which rely on weaker supervision. For instance, when using LLaMA, Coconut still achieves around 20\% accuracy on GSM8K after noise injection, compared to a clean performance of around 34\%.

These observations support our hypothesis that shortcut behavior naturally emerges in latent reasoning. Unlike explicit CoT, which enforces token-level supervision and tightly aligns reasoning steps to textual representations, latent reasoning lacks direct constraints on intermediate states, making it easier for models to bypass structured reasoning. Moreover, this phenomenon is particularly pronounced in methods trained with weaker supervision, where learning signals are primarily outcome-level and intermediate states receive only weak supervision.

\subsection{Investigation with Attention Score}
To further diagnose shortcut behaviors, in this subsection, we investigate token-level attention scores during final-answer generation.

\begin{figure*}[h]
\centering
\vspace{-0.1in}
    \includegraphics[width=0.85\textwidth]{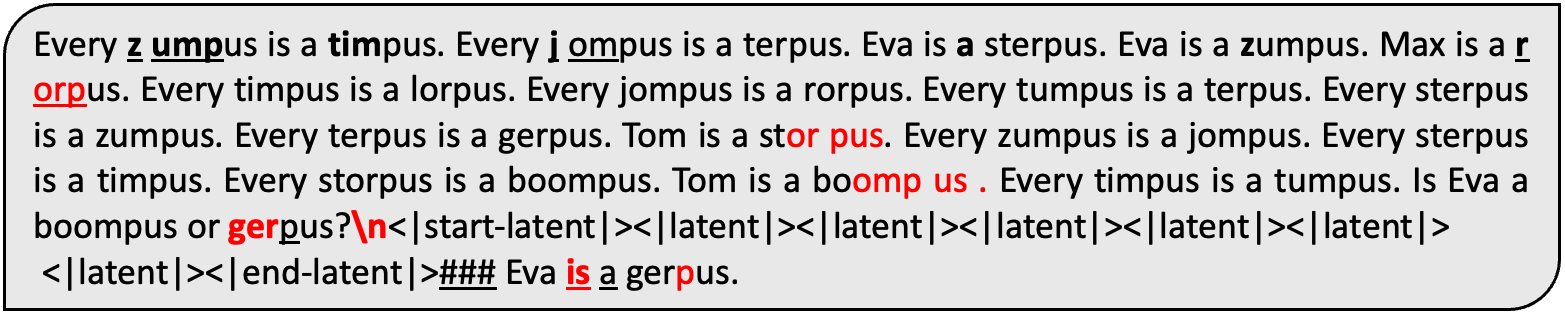}
    \caption{Tokens with Top-10 Attention Score for an Example in ProsQA Dataset}
    \label{fig:att_pros}
\end{figure*}

Figure~\ref{fig:att_pros} visualize the attention patterns of Coconut on ProsQA. The displayed text includes a complete example, including the input question, the latent reasoning process, and the final answer generation (starting from the marker {\#\#\#}). 
The final prediction, \texttt{gerpus}, consists of three tokens: \tokU{ger}, \tokB{p}, and \tokR{us}. For each output token, we highlight the top-10 tokens with the highest attention scores using the same visual cues (underline, bold, and red, respectively). We exclude the top-1 attended token, which consistently acts as an attention sink.

From the figure, we observe that the tokens with the top-10 attention scores for each answer token are exclusively drawn from the input question, rather than from latent reasoning tokens. This observation suggests that, in this example, the final prediction primarily relies on the input question instead of the intermediate latent step. 

\begin{figure*}[h]
\vspace{-0.1in}
    \centering
    \includegraphics[width=0.85\textwidth]{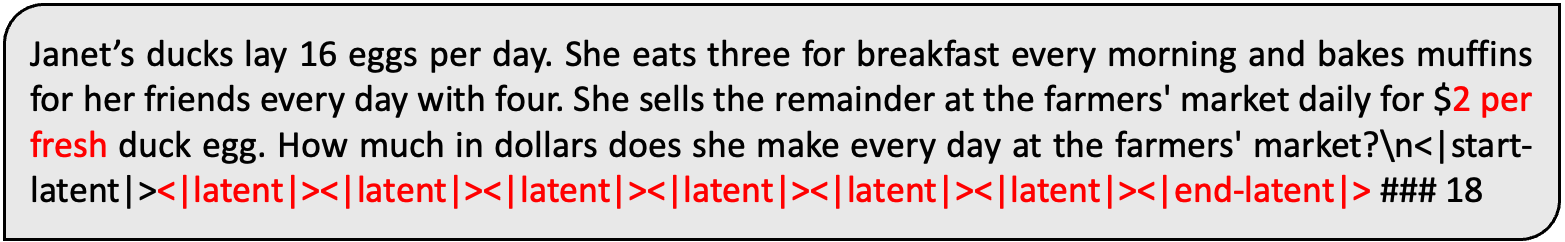}
    \caption{Tokens with Top-10 Attention Score for an Example in GSM8K Dataset}
   \label{fig:att_gsm}
    
\end{figure*}

In contrast, Figure~\ref{fig:att_gsm} presents a GSM8K example in which, when generating the final answer token \texttt{18}, the top-10 attended tokens are predominantly latent tokens, indicating a substantially different attention pattern from that observed on ProsQA.

\begin{table*}[t]
\centering
\vspace{-0.1in}
\caption{\small Performance under clean inputs vs.\ noise perturbations for different implicit/explicit reasoning methods.}
\resizebox{\linewidth}{!}{%
\begin{tabular}{l l cc cc cc cc cc}
\toprule
& & \multicolumn{2}{c}{Coconut} & \multicolumn{2}{c}{CODI} & \multicolumn{2}{c}{CoLAR} & \multicolumn{2}{c}{SIM-CoT} & \multicolumn{2}{c}{Standard CoT} \\
\cmidrule(lr){3-4} \cmidrule(lr){5-6} \cmidrule(lr){7-8} \cmidrule(lr){9-10} \cmidrule(lr){11-12}
Model & Setting & GSM8K & ProsQA & GSM8K & ProsQA & GSM8K & ProsQA & GSM8K & ProsQA & GSM8K & ProsQA \\
\midrule
Llama & clean & 36.97\% & 99.40\% & 55.57\% & 100.00\% & 25.23\% & 98.20\% & 56.03\% & 100.00\% & 61.74\% & 90.60\% \\
Llama & noise & 20.61\% & 99.40\% & 27.45\% & 99.60\% & 3.32\% & 60.92\% & 10.08\% & 97.60\% & 0.03\% & 0.40\% \\
GPT & clean & 34.09\% & 97.80\% & 43.59\% & 80.80\% & 18.44\% & 77.52\% & 42.23\% & 80.60\% & 43.56\% & 81.00\% \\
GPT & noise & 3.79\% & 89.00\%& 8.87\% & 80.80\% & 2.84\% & 46.80\% & 7.05\% & 75.60\% & 0.08\% & 0.00\% \\
\bottomrule
\end{tabular}%
}
\label{tab:noise}
\end{table*}

\section{Investigation on the BFS Mechanism}

\label{sec:bfs}
\subsection{Experiment Settings}

We focus our investigation of the parallel BFS hypothesis primarily on Coconut~\citep{hao2024training}, as it is the first implicit reasoning framework that explicitly motivate latent reasoning through the conjecture of BFS in continuous space. To minimize the influence of shortcut effects, we focus on Coconut with GPT-2 on GSM8K, which exhibits the weakest shortcut behavior.

However, in our initial experiments, we observe a collapse of Coconut's reasoning pattern during inference, which prevents a systematic investigation of BFS-like exploration. In this subsection, we first characterize this collapse behavior in detail and then introduce an improved variant of Coconut to mitigate it.


\paragraph{Collapse of the Reasoning Pattern in Coconut} 

In addition to the shortcut reliance, we observe a collapse behavior in Coconut when the number of latent steps at inference time is set to be smaller than its default configuration. Specifically, when reducing the latent length without explicitly inserting the token that indicates final-solution generation, the model often directly produces the final answer, instead of continuing to generate textual reasoning for the remaining steps. This behavior deviates from the intended training design of Coconut. During training, the model is optimized in a stage-wise manner, where latent reasoning steps are gradually increased while explicit textual steps are correspondingly reduced. Under such a scheme, when the number of latent steps is small, the model is expected to generate the remaining textual steps before producing the final solution.

This collapse behavior may arise from later training stages overriding inference behaviors learned in earlier stages. As training progresses toward stages with more latent steps and fewer textual steps, the model may gradually lose the stepwise alignment between latent states and textual predictions learned in earlier stages. Consequently, the model tends to collapse to a degenerate inference pattern, where encountering the \texttt{<|end-of-latent|>} token directly triggers final-answer generation, regardless of whether sufficient intermediate reasoning has been performed.

\paragraph{Improved Coconut}


To mitigate this issue and enable a more faithful examination of the BFS hypothesis, we modify Coconut’s training scheme by adjusting how training data are sampled across stages. Instead of training each stage with stage-wise data, we allow later stages to incorporate data from earlier stages. Specifically, at training stage $k$, the proportion of data sampled from stage $i$ ($i \leq k$) is set to be proportional to $(i+1)$, ensuring that examples with fewer latent steps remain consistently presented throughout training.
By maintaining exposure to earlier-stage training examples, this design prevents the model from forgetting the step-wise correspondence between latent and textual reasoning steps learned in previous stages. 

Empirically, we find that under the revised training scheme, the model exhibits more stable behavior when evaluated with fewer latent steps: rather than directly generating the final solution, it correctly produces the remaining reasoning steps in text space. More importantly, even under the default inference setting, where all reasoning is performed entirely in latent space, this modification leads to substantial accuracy improvements of Coconut. On GSM8K-Aug, GPT-2’s test accuracy increases from 34.09\% to 41.06\%, and on GSM8K-Aug-NL (a variant of GSM8K-Aug whose reasoning steps are described with natural language instead of mathematical formulas), accuracy improves from 24.90\% to 33.48\%. These results indicate that preserving exposure to earlier-stage training examples enables the model to learn more effective intermediate latent steps and thereby improves the overall reasoning quality. Notably, the improved training scheme, when combined with an additional slight modification to the supervision design, can help mitigate Coconut's shortcut issues in the ProsQA dataset. More discussions can be found in Appendix~\ref{appd:solve}.

\paragraph{BFS Verification via Latent–Text Hybrid Rollouts}
After mitigating the collapse behavior, we proceed to verify the BFS hypothesis by explicitly examining whether a latent step can encode multiple reasoning trajectories. To this end, we adopt a hybrid reasoning setup in which a prefix of the reasoning process is executed in latent space, followed by explicit reasoning in text space.

Specifically, we fix a given latent prefix and generate the remaining reasoning steps in text space using stochastic decoding with high temperature ($T=1$). Starting with each latent step, we perform 100 independent rollouts and analyze the diversity of both next-step predictions and the final outcomes. By varying the number of prefix latent steps, we further examine how this distribution evolves as more reasoning is encoded in the latent space.

To provide a controlled comparison, we construct a text-only baseline in which all reasoning steps are generated in text space. In this baseline, the first $n$ reasoning steps are decoded deterministically, while the remaining steps are generated using the same stochastic decoding procedure and the same number of rollouts. This design ensures that the two settings differ only in whether the initial reasoning steps are encoded in latent space or explicitly described as textual tokens.

By comparing the diversity of rollouts across these two settings, we can directly assess whether encoding reasoning steps in latent states allows the model to maintain a broader set of reasoning possibilities than explicit token-level decoding, thereby providing empirical evidence for BFS-like behavior in latent reasoning.

\begin{table*}[t]
\centering
\caption{\small Distribution of possibilities of final results with various numbers of prefix steps}
\label{tab:poss}
\resizebox{0.98\textwidth}{!}{\begin{tabular}{c c c c c| c c c c| c c c c | c c c c}
\toprule
\# prefix step 
& \multicolumn{4}{c}{Latent reasoning (next)} 
& \multicolumn{4}{c}{Latent reasoning (final)} 
& \multicolumn{4}{c}{Explicit reasoning (next)} 
& \multicolumn{4}{c}{Explicit reasoning (final)} \\
\cmidrule(lr){2-5} \cmidrule(lr){6-9} \cmidrule(lr){10-13}\cmidrule(lr){14-17}
 & avg & median & min & max  & avg & median & min & max & avg & median & min & max  & avg & median & min & max \\
\midrule
1 & 18.75 & 14 & 1& 87& 28.35 & 23 & 1 & 95 & 3.68 &2 & 1& 87 & 9.32 & 4 & 1 & 95 \\
2 & 20.38&15&1& 89&25.50 & 19 & 1 & 96 & 3.31 & 1 & 1 &65 &  6.59 & 2 & 1 & 99 \\
3 & 20.00&13&1& 97&21.82 & 13 & 1 & 96 & 2.73 & 1& 1& 80 & 4.17 & 1 & 1 & 90 \\
4 & 17.22&10&1&92 & 17.74 & 10 & 1 & 94 & 2.01 & 1 &  1 & 60 & 2.42 & 1 & 1 & 70 \\
5 & 15.84 & 9  & 1 & 91 &15.84 & 9  & 1 & 91 & 1.27 &  1 & 1 & 27 & 1.37 & 1 & 1 & 50 \\
\bottomrule
\end{tabular}}
\vspace{-0.1in}
\end{table*}

\begin{table}[t]
\centering
\caption{\small Comparison between implicit and explicit results across different prefix steps.}
\vspace{-0.1in}
\resizebox{0.50\textwidth}{!}{\begin{tabular}{c c c c c}
\toprule
& \multicolumn{2}{c}{Implicit} & \multicolumn{2}{c}{Explicit} \\
\cmidrule(lr){2-3} \cmidrule(lr){4-5}
\# prefix step  & Pass@100 & Maj@100 & Pass@100 & Maj@100 \\
\midrule
1 & 82.34\% & 44.20\% & 62.17\% & 44.12\% \\
2 & 78.62\% & 41.70\% & 55.34\% & 44.05\% \\
3 & 75.74\% & 39.95\% & 48.30\% & 42.71\% \\
4 & 70.36\% & 39.73\% & 45.87\% & 43.52\% \\
5 & 69.07\% & 39.42\% & 44.05\% & 43.59\% \\
\bottomrule
\end{tabular}}
\vspace{-0.2in}
\label{tab:pass}
\end{table}

\begin{table*}[ht]
\centering
\vspace{-0.1in}
\caption{Distribution of possibilities and performance across methods (GPT-2)}
\vspace{-0.1in}
\resizebox{0.8\textwidth}{!}{\begin{tabular}{l c c c c c c c}
\toprule
& \multicolumn{4}{c}{Distribution of possibilities} & \multicolumn{3}{c}{Performance} \\
\cmidrule(lr){2-5} \cmidrule(lr){6-8}
 & avg & median & min & max & acc (greedy) & Pass@100 & majority vote \\
\midrule
Improved Coconut     & 15.84 & 9 & 1 & 91 & 34.09\% & 69.07\% & 39.42\% \\
CODI           & 12.96 & 6 & 1 & 83 & 43.59\% & 70.43\% & 42.23\% \\
SIM-CoT  & 13.57 & 7 & 1 & 89 & 42.23\% & 69.60\% & 43.21\% \\
CoLAR  & 3.21  & 1 & 1 & 45 & 18.44\% & 23.28\% & 18.42\% \\
\bottomrule
\end{tabular}}
\label{tab:all_poss_gpt}
\vspace{-0.05in}
\end{table*}

\begin{table*}[ht]
\centering
\caption{\small Distribution of possibilities and performance across methods (\textsc{LLaMA3.2-1B}).}
\vspace{-0.1in}
\resizebox{0.8\textwidth}{!}{\begin{tabular}{l c c c c c c c}
\toprule
\textbf{llama} & \multicolumn{4}{c}{Distribution of possibilities} & \multicolumn{3}{c}{Performance} \\
\cmidrule(lr){2-5} \cmidrule(lr){6-8}
 & avg & median & min & max & acc (greedy) & majority vote & Pass@100 \\
\midrule
Improved Coconut         & 10.0 & 5 & 1 & 84 & 39.68\% & 40.21\% & 59.00\% \\
CODI           & 6.39  & 2 & 1 & 67 & 55.41\% & 55.57\% & 73.84\% \\
SIM-CoT& 7.46  & 2 & 1 & 65 & 56.01\% & 55.50\% & 72.93\% \\
CoLAR & 7.63  & 2 & 1 & 73 & 25.48\% & 25.70\% & 33.21\% \\
\bottomrule
\end{tabular}}
\label{tab:all_poss_llama}
\vspace{-0.1in}
\end{table*}

\subsection{Experiment Results}
\paragraph{Main Results} Table~\ref{tab:poss} reports the distributions of both  next-step predictions and final outcomes under latent and explicit reasoning with different numbers of prefix steps.

Compared to explicit CoT, latent CoT consistently results in greater diversity in both next-step predictions and final answers across all prefix-step settings. This observation supports the core intuition of Coconut: by avoiding decoding intermediate steps into deterministic text, latent reasoning can encode a richer set of possible reasoning paths.

However, we observe that increasing the number of latent prefix steps does not lead to a continuous expansion of the space of the subsequent step. Instead, the number of distinct answers initially increases slightly and then consistently decreases as the number of latent steps increases. This behavior indicates that latent reasoning does not exactly exhibit a BFS process. Under a true BFS mechanism, increasing the depth of latent reasoning would accumulate possibilities from earlier steps, leading to a continuous growth in the number of candidate solutions. The observed trend instead suggests that implicit pruning occurs within the latent reasoning process, limiting the diversity of surviving trajectories. This is further supported by the distribution of final outcomes, where the number of distinct solutions continues to decrease as more reasoning steps are conducted in the latent space.

To further investigate whether increased diversity at the level of final outcomes can help improve final prediction accuracy, we additionally report Pass@100 (the empirical frequency that at least one correct solution is found among the 100 sampled outputs) and majority-vote accuracy in Table~\ref{tab:pass}.
The results show that, while latent reasoning consistently achieves higher Pass@100, it does not lead to higher accuracy under majority voting. Specifically, across different numbers of prefix steps, implicit reasoning achieves Pass@100 that is consistently more than 20\% higher than that of explicit reasoning. However, its majority-vote accuracy is lower, particularly at larger prefix lengths, resulting in around 3–4\% below that of explicit reasoning.
This suggests that although latent reasoning preserves a larger set of candidate solutions, the additional diversity does not necessarily concentrate probability mass on the correct answer. This highlights that increased outcome-level diversity alone is insufficient to guarantee effective aggregation toward the correct solution in latent reasoning. We conjecture that this challenge arises from the limited capacity of the model to simultaneously maintain diverse candidate reasoning trajectories while reliably identifying and amplifying the correct one during the reasoning process.



\paragraph{Additional Results with Other Methods}
To examine whether the observed trends generalize beyond Coconut, we present additional results with other latent reasoning methods. Since some methods do not support hybrid inference with partially latent and partially textual steps, we perform all reasoning in latent space and stochastically sample the final answer 100 times and analyze the resulting outcome distributions.
The results with GPT-2 and LLaMA3.2-1B-Instruct are shown in Table~\ref{tab:all_poss_gpt} and Table~\ref{tab:all_poss_llama}, respectively.

As shown in both tables, methods with weaker supervision, such as the improved Coconut, encode the largest number of distinct solution possibilities, whereas methods with stronger supervision, such as CoLaR, exhibit the smallest  diversity. For example, with GPT-2, the average number of candidate final outcomes for the improved Coconut is 15.84, compared to only 3.21 for CoLaR.
These observations are intuitive: methods trained with weaker or more implicit supervision allow greater freedom in intermediate reasoning, enabling the model to maintain multiple plausible solution trajectories. In contrast, methods with stronger supervision impose tighter constraints on intermediate representations, which naturally reduces solution diversity. 

Notably, although CODI employs weaker supervision than CoLaR and SIM-CoT, it exhibits a smaller number of distinct solution possibilities. This might be because CODI contains stronger shortcut signals than the other two methods, biasing the model toward directly mapping inputs to final answers and thus limiting the exploration of alternative reasoning trajectories.

Additionally, with the exception of CoLaR, which combines strong supervision with limited solution diversity, all other methods achieve substantially higher Pass@100 than their greedy decoding accuracy. However, their majority-vote performance remains comparable to, or in some cases lower than, greedy decoding. For instance, with GPT-2, the Pass@100 for both SIM-CoT and CODI are around 70\%, but the accuracies of majority vote are only around 43\%.
This indicates that the mismatch between increased solution diversity and effective aggregation is not unique to Coconut but common in other latent reasoning methods.

\paragraph{Trade-off between supervision strength and latent exploration}
Based on our investigation of shortcut behavior and BFS conjecture, we observe a trade-off associated with the strength of supervision. While stronger supervision reduces the tendency toward shortcut behavior by constraining latent representations more tightly, it limits the capacity of latent states to
encode more plausible trajectories. Conversely, relaxing supervision increases representational flexibility but risks shortcut behaviors. Therefore, our investigations highlight the need for future work to better balance this trade-off and to develop methods that can reliably identify correct solutions from multiple possibilities encoded in latent reasoning. 

\section{Conclusion}
In our work, we present a comprehensive analysis of current latent reasoning models and identify several key limitations. Through a set of analytical experiments, we find that many existing methods exhibit varying degrees of shortcut behavior, achieving high performance without meaningfully relying on intermediate latent reasoning. Moreover, while latent representations can encode multiple possibilities, they do not faithfully support BFS-like exploration, but instead exhibit implicit pruning and selection. Together, these observations suggest a fundamental trade-off associated with supervision strength, where increasing supervision reduces shortcut behavior but limits latent exploration, highlighting the need for future methods that better balance these two aspects.





\bibliography{custom}

\appendix
\section{Additional Experimental Setup}\label{appd:exp_setup}
For the experiments in Section 4, we use author-released checkpoints for CODI, SIM-CoT, and CoLaR on GSM8K whenever available. All remaining settings require training from scratch. Details of the hyperparameters can be found as follows:
\begin{itemize}
    \item \textbf{Coconut.} For GSM8K, we use the AdamW optimizer with a learning rate of 1e-4, weight decay of 0.01, and train for 25 epochs. The maximum number of stages is set to 3, with 3 epochs per stage. For ProsQA, we use the same optimizer and learning rate, train for 50 epochs, set the maximum number of stages to 6, and use 5 epochs per stage.
    \item \textbf{CODI.} We use the AdamW optimizer with a learning rate of 1e-3 for GPT-2 and 2e-4 for LLaMA, weight decay 0.01, and train for 20 epochs for both models. For both models, the number of latent steps is set to 6, and LoRA is applied with rank 128 and scaling factor $\alpha=32$.
    \item \textbf{SIM-CoT.} We use the AdamW optimizer with a learning rate of 1e-3 for GPT-2 and 2e-4 for LLaMA, weight decay 0.01, and train for 40 epochs for both models. For both models, the number of latent steps is set to 6, and LoRA is applied with rank 128 and scaling factor $\alpha=32$.
    \item \textbf{CoLaR.} On ProsQA, we perform full fine-tuning for GPT-2 and LoRA-based fine-tuning for LLaMA (rank 128, $\alpha=32$). For both models, we use AdamW with a learning rate of 1e-4, weight decay of 0.01, and train for 50 epochs. The maximum compression factor is set to 5.
\end{itemize}

\section{Effects of Improved Coconut in Mitigating Shortcut}\label{appd:solve}
In this section, we provide some empirical results showing that the improved Coconut training scheme we introduced in Section can help mitigate the shortcut issue of Coconut in the ProsQA dataset. 

Notably, in addition to the improved training scheme, which reuses data from the previous stages in the later stages, an additional modification of the training scheme is required to mitigate the shortcut issue in datasets with structured reasoning, like ProsQA. Specifically, in the early training stages, when conditioning on previous latent steps, supervision is applied only to the
next-step prediction, rather than to the remaining reasoning steps or the final answer. The intuition behind this design is to avoid consistently exposing the final solution during training, especially in the later stages of the training, thereby
preventing the model from learning a direct question-to-answer mapping and encouraging stepwise reasoning.

\begin{table*}[ht]
\centering
\caption{Performance under clean and noisy latent for different  variants of Coconut.}\label{tab:abl}
\begin{tabular}{l c c c c}
\toprule
 & Coconut 
 & + Cross-stage data reuse 
 & + Next-step-only supervision 
 & + Both \\
\midrule
Clean & 97.80\% & 97.40\% & 93.60\% & 93.60\% \\
Noise & 97.80\% & 94.80\% & 93.80\% & 4.40\% \\
\bottomrule
\end{tabular}
\label{tab:noise-ablation}
\end{table*}

To demonstrate the mitigation of shortcut behavior, Table~\ref{tab:abl} reports performance under latent noise injection, together with ablation results for different variants of the proposed design choices.

As shown in the table, when both components, including the cross-stage data reuse and next-step-only supervision in early training stages, are applied, model performance drops to around 4\% under latent noise perturbations. This sharp degradation indicates that shortcut reliance has been effectively mitigated, as disrupting latent representations now severely impairs performance.
In contrast, when only one of the two components is applied, the model retains relatively high accuracy under noise, suggesting that shortcut behavior is still present. These results indicate that both design choices are necessary to effectively eliminate shortcut reliance in Coconut.

\end{document}